\pdfoutput=1
\documentclass[10pt,twocolumn,letterpaper]{article}

\usepackage[pagenumbers]{iccv} 

\usepackage{multirow}
\usepackage{tcolorbox}

\definecolor{iccvblue}{rgb}{0.21,0.49,0.74}
\definecolor{highlightcolor}{RGB}{0,134,139}
\definecolor{demphgray}{RGB}{144,144,144}
\usepackage[pagebackref,breaklinks,colorlinks,allcolors=iccvblue]{hyperref}

\def\confName{ICCV}
\def\confYear{2025}

\title{FedMLLM: Federated Fine-tuning MLLM on Multimodal Heterogeneity Data}

\author{
Binqian~Xu\textsuperscript{1} \quad Xiangbo~Shu\textsuperscript{1,*} \quad Haiyang~Mei\textsuperscript{2} \quad Guosen~Xie\textsuperscript{1} \quad
Basura~Fernando\textsuperscript{3} \quad
Jinhui~Tang\textsuperscript{1} \\
\textsuperscript{1}Nanjing University of Science and Technology
\textsuperscript{2}Show Lab, National University of Singapore \\
\textsuperscript{3}Institute of High-Performance Computing, A*STAR \\
}

\begin{document}
\maketitle

\footnotetext{Corresponding author}

\begin{abstract}
Multimodal Large Language Models (MLLMs) have made significant advancements, demonstrating powerful capabilities in processing and understanding multimodal data. Fine-tuning MLLMs with Federated Learning (FL) allows for expanding the training data scope by including private data sources, thereby enhancing their practical applicability in privacy-sensitive domains. However, current research remains in the early stage, particularly in addressing the \textbf{multimodal heterogeneities} in real-world applications. In this paper, we introduce a benchmark to evaluate the performance of federated fine-tuning of MLLMs across various multimodal heterogeneous scenarios, laying the groundwork for future research in the field. Our benchmark includes two lightweight MLLMs, two downstream tasks, three evaluation metrics, and five datasets across three domains, along with six comparison baselines, covering over ten types of modality heterogeneities across four multimodal scenarios. To address the challenges posed by multimodal heterogeneity, we develop a general FedMLLM framework that integrates classic FL methods alongside two modality-agnostic strategies. Extensive experimental results show that our proposed FL paradigm improves the performance of MLLMs by broadening the range of training data and mitigating multimodal heterogeneity. Code is available in supplementary materials.
\end{abstract}

\section{Introduction}
\label{sec:intro}
Multimodal Large Language Models (MLLMs), which use Large Language Models (LLMs) as core components for various multimodal tasks, have rapidly emerged as a research hotspot~\cite{yin2023survey}. The exceptional performance of MLLMs and their potential as a pathway to artificial general intelligence have driven this trend. 
However, the decreasing amount of available public data~\cite{villalobos2024position} is an urgent issue that hinders the further development and application of MLLMs in the real world~\cite{yao2024minicpm, liu2023visual}. To expand the available data scope, a practical solution is to fine-tune existing MLLMs, trained on public data, with unused private data via Federated Learning (FL)~\cite{mcmahan2017communication, kairouz2021advances}.

\begin{figure}
    \centering
    \includegraphics[width=0.95\linewidth]{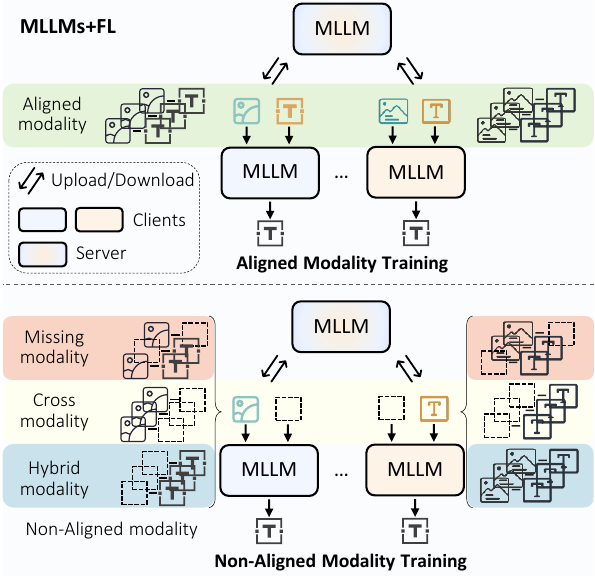}
    \caption{MLLMs$+$FL training on decentralized multimodal data, including both aligned and non-aligned modality training, where non-aligned modality data across clients contain multimodal heterogeneity compared to aligned modality data.}
    \label{fig:idea}
    \vspace{-5pt}
\end{figure}

\begin{figure*}[!t]
    \centering
    \includegraphics[width=0.9\linewidth]{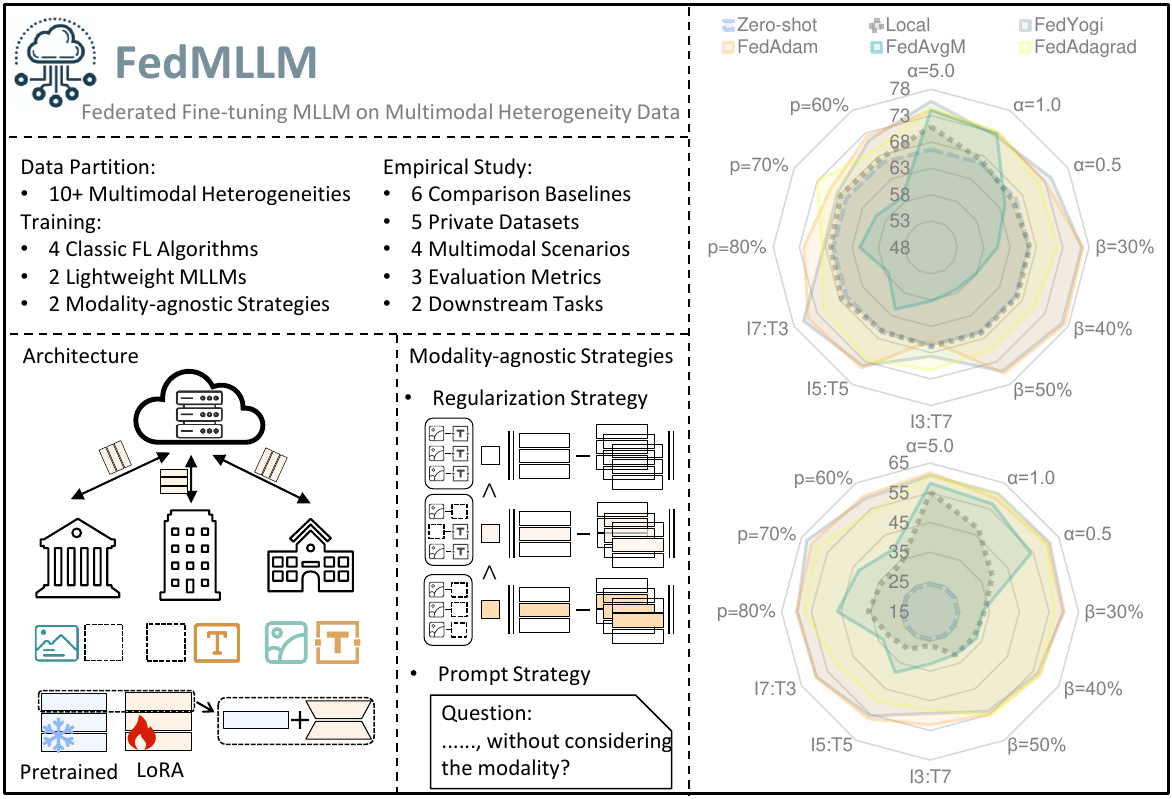}
    \caption{Overview of FedMLLM and its performance show. FedMLLM has fully deployed over ten multimodal heterogeneities, four classic FL algorithms, two lightweight MLLMs, two modality-agnostic strategies, and supports six comparison baselines, 
    five private datasets,
    four multimodal scenarios,
    three evaluation metrics, and two downstream tasks. The results of the multimodal scenarios come from federated fine-tuning of MLLM on the Hateful-Memes (top) and CrisisMMD (bottom) datasets across twelve multimodal scenarios.}
    \label{fig:fedmllm}
\end{figure*}

\begin{figure*}[!t]
    \centering
    \includegraphics[width=0.95\linewidth]{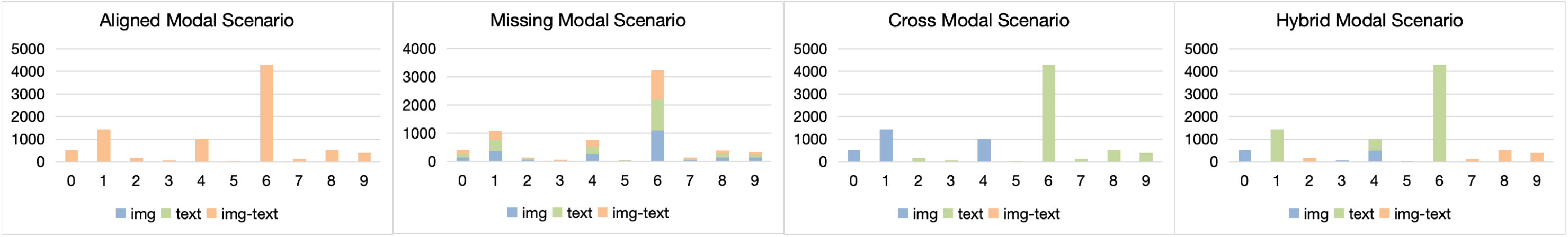}
    \caption{Visualization of modality counts and types across four multimodal scenarios on the Hateful-Memes dataset. The horizontal axis represents clients, and the vertical axis shows the sample count. Differences in modality across clients illustrate multimodal heterogeneity.}
    \label{fig:non-aligned_modal}
    \vspace{-5pt}
\end{figure*}

\begin{figure*}
    \centering
    \includegraphics[width=0.95\linewidth]{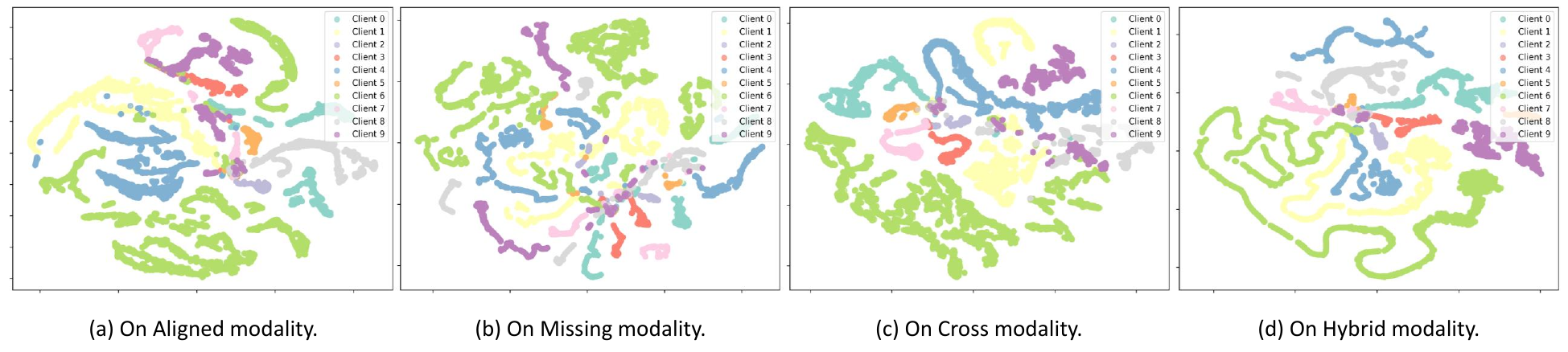}
    \caption{The t-SNE visualization of embeddings across clients from multimodal scenarios, with different colors denoting various clients.}
    \label{fig:tsne-crisis}
\end{figure*}

\begin{figure*}[!t]
    \centering
    \includegraphics[width=0.95\linewidth]{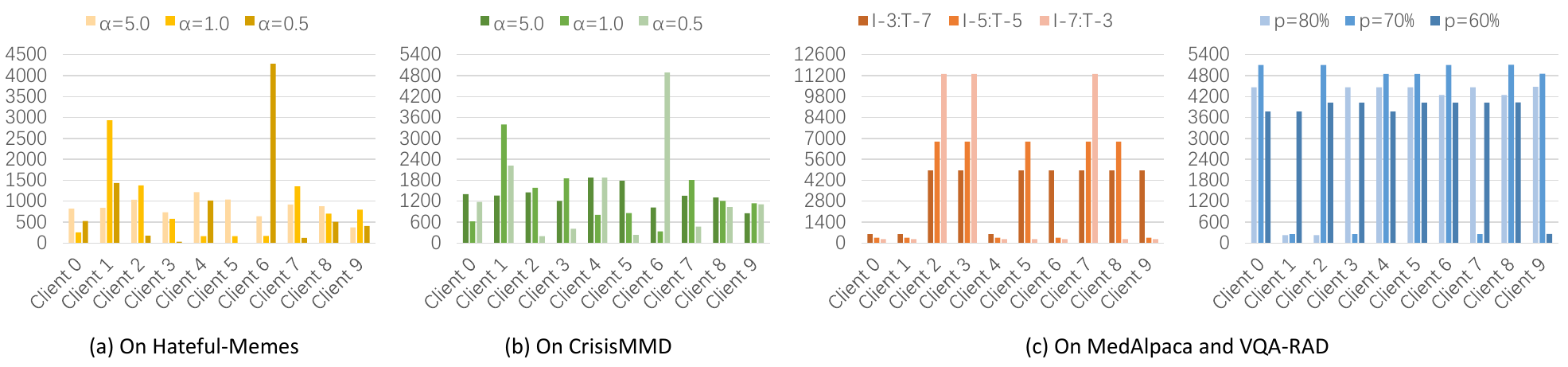}
    \caption{Visualization of sample counts across four multimodal scenarios in the four datasets. Data volume differences across clients exacerbate heterogeneity challenges. Particularly the Cross and Hybrid Modal scenarios constructed by MedAlpaca and VQA-RAD.}
    \label{fig:sample_count}
    \vspace{-5pt}
\end{figure*}

However, there is limited research on fine-tuning MLLMs under FL~\cite{zhang2024mllm,xiong2025pilot}.
There are two possible reasons. First, the large parameter size of MLLMs~\cite{wu2023multimodal} demands significant computational resources for fine-tuning, while FL typically assumes clients have limited resources~\cite{li2020review}. Excitingly, recent solutions, such as lightweight MLLMs~\cite{yao2024minicpm, li2024mini}, PEFT methods~\cite{hu2021lora, lester2021power}, and model quantization~\cite{lin2024awq}, offer support for training MLLMs in FL environments. 
In this work, we successfully fine-tune MLLMs on the client side of FL, using quantized, lightweight MLLMs~\cite{yao2024minicpm} and training them via LoRA~\cite{hu2021lora} on a single GPU. This approach is affordable in cross-silo FL settings~\cite{huang2022cross}, but cross-device FL~\cite{mammen2021federated} remains unfeasible and requires further exploration in future work.

Another possible reason is that current research has not focused on the different challenges faced by MLLMs$+$FL compared to LLMs$+$FL, and thus most studies have concentrated on LLMs$+$FL~\cite{ye2024openfedllm, kuang2024federatedscope, zhang2024towards}. Compared to LLMs$+$FL, MLLMs$+$FL supports more modal inputs, making them more applicable to real-world privacy-preserve scenarios~\cite{yin2023survey}. For example, in natural disasters or emergencies (e.g., earthquakes, floods), integrating multimodal data such as images or videos (on-site photos), text (social media reports), and audio (distress signals) is essential for quickly analyzing the situation~\cite{alam2018crisismmd}. 
Additionally, in hospitals~\cite{lau2018dataset}, besides text-based medical records, CT images are available for doctors to comprehensively analyze a patient's condition—an ability that LLMs, which process only text, cannot achieve. Meanwhile, real-world data is influenced by various conditions and situations, making it unrealistic for all clients to have complete modality samples. The absence of certain modalities in samples across clients introduces a new challenge in MLLMs$+$FL, \ie, multimodal heterogeneity.

To comprehensively explore heterogeneity in multimodal data, especially multimodal heterogeneity, we construct four modality scenarios based on existing datasets. In~\cref{fig:idea}, multimodal data can be categorized into aligned modality~\cite{feng2023fedmultimodal} and non-aligned modality (i.e., missing modality~\cite{ma2021smil, che2024leveraging}, cross modality~\cite{yang2024cross}, and hybrid modality~\cite{yumultimodal, chen2022fedmsplit}) based on the modality distribution of samples. When federated fine-tuning MLLMs on aligned modality data, client data heterogeneity resembles that in LLM fine-tuning with single-modal text data. However, training MLLMs on non-aligned modality data introduces multimodal heterogeneity. To intuitively demonstrate the heterogeneity in multimodal data, as shown in~\cref{fig:non-aligned_modal,fig:tsne-crisis,fig:sample_count}, we conduct a visual analysis from multiple perspectives on the multimodal scenarios constructed from different datasets. Among them, visualizations of modality count, types, and feature levels further emphasize the presence of multimodal heterogeneity. More visualizations are in Suppl.

Building on the above multimodal scenarios, we propose a benchmark for federated fine-tuning MLLMs. Specifically, four multimodal scenarios covering more than ten types of modality heterogeneity across five datasets are first constructed. These datasets include classification and medical visual question-answer tasks spanning the domains of social media, natural disasters, and medical knowledge.
A general FedMLLM framework based on LoRA is then presented for training and testing on these multimodal scenarios. Using FedMLLM, we compare six baselines and two lightweight MLLMs across three evaluation metrics: AUC, F1, and Accuracy. Additionally, we address the bias introduced by multimodal heterogeneity using two simple yet effective modal-agnostic strategies: improved prompts and regularization techniques. The overview of FedMLLM is as illustrated in~\cref{fig:fedmllm}.

Overall, our contributions can be summarized in three folds. 1) We explore the fine-tuning MLLM benchmark on decentralized multimodal data to better adapt to downstream tasks in comprehensive multimodal federated scenarios, which highlights a potential development direction for MLLMs. 2) We build a general FedMLLM framework to better address the challenges of multimodal heterogeneity by elegantly incorporating two modality-agnostic strategies on top of multiple FL baselines. 3) We conduct extensive experiments, related to 5 downstream datasets, 6 comparison baselines, and 4 multimodal scenarios, to show that FL paradigm enhances MLLM performance by broadening the training data scope without bells and whistles.

\section{Multimodal Scenarios Construction}
\label{sec:multimodal}
For a comprehensive simulation of multimodal scenarios in the context of FL, we refer to~\cite{feng2023fedmultimodal,che2024leveraging,yang2024cross}. As shown in~\cref{fig:idea}, the simulation is divided into four scenarios: Aligned Modal Scenario, Missing Modal Scenario, Cross Modal Scenario, and Hybrid Modal Scenario. 

\noindent\textbf{Aligned Modal Scenario.} The dataset is partitioned using a Dirichlet distribution, similar to the traditional method~\cite{feng2023fedmultimodal}, with coefficients $\alpha=\{5.0, 1.0, 0.5\}$ controlling sample distribution among clients. A smaller coefficient increases differences in sample counts across clients. For classification tasks, sampling is based on class labels; for non-classification tasks, it depends on data types, user IDs, etc. Each client’s sample includes all modalities.

\noindent\textbf{Missing Modal Scenario.} Building on the Aligned Modal Scenario, a uniform distribution is applied to each modality per sample with a missing rate of $\beta(\%)=\{30, 40, 50\}$ to determine modality removal, resulting in samples with different missing modalities across clients. Both the Aligned and Missing Modal Scenarios require the constructed dataset to have complete modalities.

\noindent\textbf{Cross Modal Scenario.} This scenario involves two approaches: one for a full-modal dataset (Hateful-Memes, CrisisMMD), where samples are allocated to clients using a Dirichlet distribution, followed by random assignment of image-only and text-only clients in proportion. The other approach is for datasets with only the image modality (VQA-RAD) or text modality (MedAlpaca), where clients are randomly assigned in proportion and the dataset is evenly distributed to the corresponding clients. The proportions used in this study are (\{$\text{I-}3:\text{T-}7$, $\text{I-}5:\text{T-}5$, $\text{I-}7:\text{T-}3$\}), where I and T denote image-only and text-only clients.

\noindent\textbf{Hybrid Modal Scenario.} Each modality per client follows a Bernoulli distribution with retention probabilities $p(\%)=\{80, 70, 60\}$. This approach applies to both full-modal and single-modal datasets. After partitioning, clients may have either single-modal or multi-modal data. If both modalities are missing on the client side, one modality is randomly retained with a 50\% probability.

\section{FedMLLM Framework}
\subsection{Overview of FedMLLM}
As shown in the architecture part of~\cref{fig:fedmllm}, the proposed FedMLLM is divided into two parts: the cloud (server-side) and the terminal (client-side). Both parts use the same MLLM~\cite{yao2024minicpm} and share a homogeneous model architecture. In FedMLLM, we address an optimization problem similar to that of traditional FL~\cite{mcmahan2017communication}, as follows,
\begin{equation}
    \min_{w\in\mathbb{R}^d}f(w)=\frac{1}{K}\sum_{k=1}^KF_k(w)
\end{equation}
where $F_k(w)$ represents the loss function of  client $k$ among a total of $K$ clients, specifically given by $F_k(w)=\mathbb{E}_{x\sim \mathcal{D}_k}[f_k(w,x)]$, with $\mathcal{D}_k$ denoting the data distribution for the $k^{\text{th}}$ client. Here, the model parameters are given by $w=w_0+\Delta w=w_0+BA$, where $w_0$ refers to the frozen pretrained weights, and $BA$ is the trained LoRA component~\cite{hu2021lora} of the LLM.

The FL process of FedMLLM starts on the server side, which sends either the initialized LoRA weights (in the first round) or the aggregated weights (in subsequent rounds) to the selected clients to initialize their LoRA weights. Then,  client $k$ fine-tunes the initialized LoRA weights on the local dataset $\mathcal{P}_k=\{(\mathcal{I}^i,\mathcal{M}^i,\mathcal{R}^i)\}_i^{n_k}$, where $\mathcal{I}^i$, $\mathcal{M}^i$, $\mathcal{R}^i$ denote the instruction token sequences, multimodal input token sequences, and ground truth response token sequences, respectively, with $n_k=|\mathcal{P}_k|$. The loss function~\cite{yin2023survey} maximizes the probability of the current input sequence generating the next correct token, as follows,
\begin{equation}
    \mathcal{L}_{mllm}=-\log\prod_{l=1}^{L^i}p(\mathcal{R}_l^i|\mathcal{M}^i,\mathcal{I}^i,\mathcal{R}^i_{<l};w^{(t,r)}_k)
\end{equation}
where $L^i$ is the length of $\mathcal{R}^i$, and $w^{(t,r)}_k$ represents the model parameters of client $k$ at epoch $r$ of round $t$. Once local training is complete, the updated LoRA weights from the selected clients are uploaded to the server for aggregation. At this stage, one training round has been completed, and the FL training process needs to be repeated for several more rounds until the model converges.

\begin{table*}[!t]
    \centering
    \renewcommand{\arraystretch}{1.2}
    \setlength{\tabcolsep}{5pt}
    \fontsize{9}{9}\selectfont
    \begin{tabular}{l|cccccc}
    \hline
       Dataset & Domain & Evaluate metric & Task & Modality & Training set & Testing set  \\ \hline
        Hateful-Memes~\cite{kiela2020hateful} & Social media & AUC & Classification & Image\&Text & 8,500 & 1,000 \\
        CrisisMMD~\cite{alam2018crisismmd} & Natural disasters & F1 & Classification & Image\&Text & 13,608 & 2,238 \\
        MedAlpaca~\cite{han2023medalpaca} & Medical knowledge & -- & Med-QA & Text & 33,951 & -- \\
        VQA-RAD~\cite{lau2018dataset} & Medical knowledge & Accuracy & Med-VQA & Image & 1,797 & 451 \\
        SLAKE~\cite{liu2021slake} & Medical knowledge & Accuracy & Med-VQA & Image & -- & 2,094 \\ \hline
    \end{tabular}
    \caption{Summary of the five datasets included in FedMLLM. MedAlpaca serves solely for training, while SLAKE, an unseen dataset, is designated for zero-shot testing. Accuracy is evaluated by GPT-4~\cite{achiam2023gpt} on open- and closed-ended answer types in Med-VQA.}
    \label{tab:datasets}
\end{table*}

\begin{table*}[!t]
    \centering
    \renewcommand{\arraystretch}{1.2}
    \setlength{\tabcolsep}{5.5pt}
    \fontsize{9}{9}\selectfont
    \begin{tabular}{l|cc|cc|cc}
         \hline
         \multirow{2}{*}{Method} & \multicolumn{2}{c|}{$\alpha=5.0$} & \multicolumn{2}{c|}{$\alpha=1.0$} & \multicolumn{2}{c}{$\alpha=0.5$} \\ \cline{2-7}
         & Hateful-Memes & CrisisMMD & Hateful-Memes & CrisisMMD & Hateful-Memes & CrisisMMD \\ \hline
         Zero-shot & 66.57$\pm$0.00 & 24.20$\pm$0.00 & 66.57$\pm$0.00 & 24.20$\pm$0.00 & 66.57$\pm$0.00 & 24.20$\pm$0.00 \\
         Local & 70.93$\pm$0.68 & 55.27$\pm$0.29 & 67.10$\pm$0.57 & 47.34$\pm$1.56 & 65.91$\pm$1.27 & 39.12$\pm$1.45 \\
         FedYogi & 75.74$\pm$0.88 & 60.95$\pm$1.06 & 72.50$\pm$1.46 & 60.82$\pm$0.58 & 74.32$\pm$0.82 & 61.46$\pm$0.95 \\
         FedAdam & 74.02$\pm$1.42 & 61.80$\pm$0.75 & 73.24$\pm$2.43 & 59.12$\pm$0.41 & 72.97$\pm$1.23 & 60.56$\pm$0.56 \\
         FedAvgM & 74.08$\pm$1.35 & 58.28$\pm$1.77 & 72.94$\pm$1.61 & 56.87$\pm$0.11 & 64.22$\pm$1.71 & 54.22$\pm$0.42 \\
         FedAdagrad & 74.53$\pm$0.55 & 60.77$\pm$0.12 & 73.34$\pm$1.99 & 60.43$\pm$0.23 & 71.80$\pm$0.89 & 60.21$\pm$0.60 \\
         \hline
    \end{tabular}
    \caption{On Aligned Modal Scenario. $\alpha$ is the Dirichlet coefficient that controls the degree of traditional data heterogeneity.}
    \label{tab:aligned}
\end{table*}

\begin{table*}[!t]
    \centering
    \renewcommand{\arraystretch}{1.2}
    \setlength{\tabcolsep}{5.5pt}
    \fontsize{9}{9}\selectfont
    \begin{tabular}{l|cc|cc|cc}
         \hline
         \multirow{2}{*}{Method} & \multicolumn{2}{c|}{$\beta=30\%$} & \multicolumn{2}{c|}{$\beta=40\%$} & \multicolumn{2}{c}{$\beta=50\%$}  \\ \cline{2-7}
         & Hateful-Memes & CrisisMMD & Hateful-Memes & CrisisMMD & Hateful-Memes & CrisisMMD \\ \hline
         Zero-shot & 66.57$\pm$0.00 & 24.20$\pm$0.00 & 66.57$\pm$0.00 & 24.20$\pm$0.00 & 66.57$\pm$0.00 & 24.20$\pm$0.00 \\
         Local & 66.72$\pm$0.11 & 33.27$\pm$0.08 & 66.84$\pm$0.10 & 31.99$\pm$0.11 & 66.80$\pm$0.11 & 31.78$\pm$0.22 \\
         FedYogi & 76.58$\pm$0.90 & 60.07$\pm$0.01 & 76.75$\pm$0.55 & 56.76$\pm$1.70 & 75.12$\pm$0.70 & 53.78$\pm$0.80 \\
         FedAdam & 76.82$\pm$0.05 & 59.54$\pm$1.07 & 77.28$\pm$0.52 & 57.64$\pm$0.32 & 75.58$\pm$1.72 & 54.82$\pm$0.49 \\
         FedAvgM & 60.62$\pm$2.45 & 32.64$\pm$0.53 & 57.90$\pm$1.46 & 32.97$\pm$1.61 & 57.34$\pm$1.35 & 31.72$\pm$1.27 \\
         FedAdagrad & 71.97$\pm$0.56 & 56.96$\pm$0.31 & 70.64$\pm$0.32 & 57.48$\pm$1.18 & 70.59$\pm$0.27 & 55.55$\pm$0.63\\
         \hline
    \end{tabular}
    \caption{On Missing Modal Scenario. For example, $\beta=50\%$ means each modality is 50\% likely to be missing per sample.}
    \label{tab:missing}
    \vspace{-5pt}
\end{table*}

\subsection{Modality-agnostic Strategies}

\noindent\textbf{Prompt Strategy.} In decentralized multimodal data scenarios, different modality conditions among clients further complicate the data heterogeneity problem. Lacking modalities may introduce additional modality bias into the model's learning process, affecting the convergence of the global model. A common solution is to address this issue at the data source by using a large-scale pre-trained model to supplement the modality~\cite{che2024leveraging}. Nevertheless, it often incurs additional costs. 
To avoid these costs, we aim to mitigate the impact of modality changes on task outcomes by explicitly prompting model training. Inspired by this, we augment the original prompt (\textit{Original prompt: }Is the content hateful based on the text and image?) by introducing a modality-agnostic prompt (\textit{Improved prompt: }Is the content hateful, without considering the modality?) strategy to enhance the model's robustness and generalization ability. More detailed prompt templates can be found in Suppl.

\noindent\textbf{Regularization Strategy.} The regularization term originally introduced by FedProx~\cite{li2020federated} constrains the direction of local updates toward the global model, serving as a general method for mitigating heterogeneity. However, the average aggregation strategies used in FedProx and FedAvg~\cite{mcmahan2017communication} can exacerbate instability during federated fine-tuning of MLLM, leading to a gradual increase in loss and its variance (see in Suppl.). In contrast, FedAvgM employs server momentum~\cite{hsu2019measuring}, while FedAdam, FedYogi, and FedAdagrad use adaptive optimizers~\cite{reddiadaptive}, allowing them to handle instability more effectively.

To reduce the negative effect of modal heterogeneity, a straightforward approach is to integrate the FedProx regularization term into the stable FL baselines. This traditional regularization strategy with a fixed coefficient is not suitable for various non-aligned modal scenarios. Specifically, as the issue of missing modalities becomes more severe in such scenarios, it is more reasonable to strengthen the regularization to further prevent overfitting to a particular modality. At the same time, to preserve both the low-level modality-specific and high-level task-specific information, we only regularize the intermediate modality-agnostic layers. Therefore, an adaptive regularization strategy is presented, as follows, 
\begin{equation}
\begin{aligned}
    \mathcal{L} &= \mathcal{L}_{mllm} + \gamma\cdot\mathcal{A}\cdot\|\Delta\tilde{w}-\Delta w_k\|^2 \\
    \mathcal{A} &= [\underbrace{0,\dots,0}_{S},1,\dots,1, \underbrace{0,\dots,0}_{S}]^\top
\end{aligned}
\end{equation}
where $\gamma$ is a hyperparameter related to the degree of modality missing, $\mathcal{A}$ is the layer-wise mask, $\top$ is the transpose operation, $S$ is a layer-count hyperparameter, and $\Delta\tilde{w}$ represents the server-side aggregated LoRA weight. More details are in Suppl.

\begin{table*}[!t]
    \centering
    \renewcommand{\arraystretch}{1.2}
    \setlength{\tabcolsep}{5.5pt}
    \fontsize{9}{9}\selectfont
    \begin{tabular}{l|cc|cc|cc}
         \hline
         \multirow{2}{*}{Method} & \multicolumn{2}{c|}{$\text{I-}3:\text{T-}7$} & \multicolumn{2}{c|}{$\text{I-}5:\text{T-}5$} & \multicolumn{2}{c}{$\text{I-}7:\text{T-}3$}  \\ \cline{2-7}
         & Hateful-Memes & CrisisMMD & Hateful-Memes & CrisisMMD & Hateful-Memes & CrisisMMD \\ \hline
         Zero-shot & 66.57$\pm$0.00 & 24.20$\pm$0.00 & 66.57$\pm$0.00 & 24.20$\pm$0.00 & 66.57$\pm$0.00 & 24.20$\pm$0.00 \\
         Local & 66.85$\pm$0.20 & 26.36$\pm$0.43 & 66.39$\pm$0.01 & 29.59$\pm$0.77 & 67.57$\pm$0.20 & 31.14$\pm$0.29 \\
         FedYogi & 68.71$\pm$1.69 & 49.30$\pm$1.75 & 73.93$\pm$1.16 & 55.53$\pm$2.50 & 75.94$\pm$0.08 & 59.20$\pm$1.77 \\
         FedAdam & 65.71$\pm$1.64 & 53.17$\pm$0.96 & 74.44$\pm$0.34 & 56.62$\pm$1.58 & 74.95$\pm$0.39 & 59.22$\pm$1.69 \\
         FedAvgM & 58.20$\pm$2.36 & 32.61$\pm$1.30 & 61.57$\pm$2.34 & 38.62$\pm$1.66 & 57.40$\pm$1.37 & 33.46$\pm$0.05 \\
         FedAdagrad & 71.26$\pm$1.93 & 48.02$\pm$0.87 & 72.48$\pm$0.05 & 49.97$\pm$0.90 & 71.57$\pm$0.77 & 51.13$\pm$0.42 \\
         \hline
    \end{tabular}
    \caption{On Cross Modal Scenario. For example, I-3:T-7 means three clients have only image data, and seven have only text data.}
    \label{tab:Cross}
\end{table*}

\begin{table*}[!t]
    \centering
    \renewcommand{\arraystretch}{1.2}
    \setlength{\tabcolsep}{5.5pt}
    \fontsize{9}{9}\selectfont
    \begin{tabular}{l|cc|cc|cc}
         \hline
         \multirow{2}{*}{Method} & \multicolumn{2}{c|}{$p=80\%$} & \multicolumn{2}{c|}{$p=70\%$} & \multicolumn{2}{c}{$p=60\%$}  \\ \cline{2-7}
         & Hateful-Memes & CrisisMMD & Hateful-Memes & CrisisMMD & Hateful-Memes & CrisisMMD \\ \hline
         Zero-shot & 66.57$\pm$0.00 & 24.20$\pm$0.00 & 66.57$\pm$0.00 & 24.20$\pm$0.00 & 66.57$\pm$0.00 & 24.20$\pm$0.00 \\
         Local & 67.05$\pm$0.31 & 35.84$\pm$0.23 & 67.82$\pm$0.16 & 34.00$\pm$0.42 & 67.44$\pm$0.19 & 34.18$\pm$0.20 \\
         FedYogi & 68.41$\pm$1.65 & 59.73$\pm$0.96 & 68.55$\pm$0.71 & 63.04$\pm$1.63 & 71.52$\pm$1.13 & 58.96$\pm$0.57 \\
         FedAdam & 72.17$\pm$1.62 & 60.02$\pm$0.63 & 70.12$\pm$0.54 & 61.01$\pm$1.12 & 72.36$\pm$0.97 & 59.93$\pm$0.60  \\
         FedAvgM & 61.55$\pm$1.33 & 46.26$\pm$1.78 & 60.01$\pm$2.03 & 42.77$\pm$1.05 & 58.93$\pm$1.98 & 39.44$\pm$0.57 \\ 
         FedAdagrad & 68.77$\pm$0.63 & 57.58$\pm$0.32 & 72.96$\pm$0.54 & 58.91$\pm$1.03 & 69.26$\pm$0.47 & 53.84$\pm$0.46 \\
         \hline
    \end{tabular}
    \caption{On Hybrid Modal Scenario. For example, $p=80\%$ means that each modality is 80\% likely to be retaining per client.}
    \label{tab:Hybrid}
\end{table*}

\begin{table*}[!t]
    \centering
    \renewcommand{\arraystretch}{1.2}
    \setlength{\tabcolsep}{5.5pt}
    \fontsize{9}{9}\selectfont
    \begin{tabular}{l|c|ccc|ccc}
         \hline
         \multirow{2}{*}{Method} & \multirow{2}{*}{Scenario} & \multicolumn{3}{c|}{VQA-RAD} & \multicolumn{3}{c}{SLAKE} \\ \cline{3-8}
         & & Open & Closed & Overall & Open & Closed & Overall \\ \hline
         \multicolumn{2}{c|}{Zero-shot} & 44.69$\pm$0.00 & 65.07$\pm$0.00 & 56.98$\pm$0.00 & 65.55$\pm$0.00 & 64.11$\pm$0.00 & 64.95$\pm$0.00 \\ \hline
         Local & \multirow{2}{*}{$\text{I-}3:\text{T-}7$} & 42.74$\pm$0.09 & 69.97$\pm$0.13 & 59.16$\pm$0.05 & 58.71$\pm$0.68 & 65.83$\pm$0.48 & 61.63$\pm$0.22 \\
         FL &  & 45.25$\pm$0.56 & 68.02$\pm$1.11 & 58.98$\pm$0.89 & 53.70$\pm$0.35 & 66.15$\pm$0.24 & 58.67$\pm$0.12 \\ \hline
         Local & \multirow{2}{*}{$\text{I-}5:\text{T-}5$} & 43.86$\pm$0.43 & 70.04$\pm$0.55 & 59.64$\pm$0.32 & 58.19$\pm$0.20 & 65.19$\pm$0.37 & 60.98$\pm$0.16 \\
         FL &  & 44.42$\pm$0.84 & 71.14$\pm$0.18 & 60.53$\pm$0.22 & 52.55$\pm$0.23 & 63.04$\pm$0.60 & 56.74$\pm$0.09 \\ \hline
         Local & \multirow{2}{*}{$\text{I-}7:\text{T-}3$} & 41.90$\pm$0.29 & 71.69$\pm$0.31 & 59.87$\pm$0.15 & 59.22$\pm$0.33 & 62.32$\pm$0.27 & 60.46$\pm$0.21 \\
         FL &  & 41.90$\pm$0.75 & 73.53$\pm$0.20 & 60.98$\pm$0.31 & 52.78$\pm$0.39 & 60.41$\pm$0.54 & 55.83$\pm$0.14 \\ \hline
         Local & \multirow{2}{*}{$p=80\%$} & 41.06$\pm$0.38 & 71.87$\pm$0.16 & 59.64$\pm$0.11 & 52.23$\pm$0.35 & 67.34$\pm$0.39 & 58.26$\pm$0.28 \\
         FL &  & 44.13$\pm$0.92 & 70.96$\pm$0.99 & 60.31$\pm$0.18 & 53.74$\pm$0.83 & 64.00$\pm$1.06 & 57.83$\pm$0.47 \\ \hline
         Local & \multirow{2}{*}{$p=70\%$} & 45.25$\pm$0.33 & 69.85$\pm$0.68 & 60.09$\pm$0.31 & 53.02$\pm$0.42 & 65.67$\pm$0.57 & 58.07$\pm$0.34 \\
         FL &  & 43.02$\pm$1.67 & 72.06$\pm$2.21 & 60.54$\pm$0.67 & 54.46$\pm$0.55 & 64.60$\pm$0.71 & 58.50$\pm$0.05 \\ \hline
         Local & \multirow{2}{*}{$p=60\%$} & 46.93$\pm$0.07 & 70.96$\pm$0.13 & 61.42$\pm$0.10 & 56.36$\pm$0.26 & 63.52$\pm$0.41 & 59.22$\pm$0.28 \\
         FL &  & 46.64$\pm$1.39 & 73.90$\pm$1.10 & 63.08$\pm$0.11 & 52.43$\pm$0.75 & 64.00$\pm$2.63 & 57.05$\pm$0.59 \\
         \hline
    \end{tabular}
    \caption{Comparison of Cross and Hybrid Modal Scenarios in medical datasets. The entire MedAlpaca dataset and the training data from VQA-RAD are used for training. The testing data from VQA-RAD and SLAKE (an unseen dataset) are used for evaluation. The evaluation metric is accuracy, measured by GPT-4. ``FL" denotes FedYogi, while ``Open" and ``Closed" refer to the open- and closed-end answer types.}
    \label{tab:Medical}
    \vspace{-5pt}
\end{table*}

\section{Experiments}
\subsection{Experimental Settings}

\noindent\textbf{Baselines.} 
In different multimodal scenarios, to evaluate the performance of various FL baseline, we refer to~\cite{ye2024openfedllm,ye2025fedllm} and provide four classic FL methods as baselines. FedYogi, FedAdam, and FedAdagrad~\cite{reddiadaptive} apply the centralized Yogi~\cite{zaheer2018adaptive}, Adam~\cite{kingma2014adam}, and Adagrad~\cite{duchi2011adaptive} optimizers to the server-side aggregation in the FL framework, while FedAvgm~\cite{hsu2019measuring} uses momentum updates on the server, similar to momentum applied on top of SGD. FedAvg~\cite{mcmahan2017communication} and FedProx~\cite{li2020federated} are not included in the benchmark due to their poor convergence, as shown in Suppl.
Additionally, we consider ``Zero-shot" and the local baseline for comparison. ``Zero-shot" refers to the original MLLM's test results without federated fine-tuning, while the ``Local" baseline is the average test results of models trained independently on local data. 

\noindent{\textbf{Datasets.}} 
To comprehensively evaluate different multimodal scenarios, we conduct multimodal training on four datasets from the domains of social media, natural disasters, and medical knowledge—Hateful-Memes~\cite{kiela2020hateful}, CrisisMMD~\cite{alam2018crisismmd}, MedAlpaca~\cite{han2023medalpaca}, and VQA-RAD~\cite{lau2018dataset}. The evaluation is performed on the test sets of Hateful-Memes, CrisisMMD, VQA-RAD, and SLAKE~\cite{liu2021slake}, where SLAKE serves as an unseen dataset to assess the zero-shot performance of FedMLLM. The evaluation metrics include AUC, F1, and accuracy for open-ended, closed-ended, and overall answers, with accuracy being assessed by GPT-4~\cite{achiam2023gpt}. The dataset information is shown in~\cref{tab:datasets}. More details in Suppl.

\noindent{\textbf{Configurations.}} To adapt to resource-constrained FL scenarios, we chose the existing lightweight MLLM (\eg, MiniCPM-V-2\_6-int4~\cite{yao2024minicpm}) for fine-tuning via LoRA. The LoRA parameters are set to a rank of 8, alpha of 8, and dropout of 0.05. The maximum sequence length for the model is 1200. Formatting instructions can be found in Suppl. In the FL setting, the proposed FedMLLM focuses on the cross-silo scenario~\cite{huang2022cross}, where, by default, 2 clients are randomly selected per round from total 10 clients. We also experimented with different numbers of clients. In the training configuration, all datasets start with an initial learning rate of 2e-5 and are updated using a cosine scheduler with a 0.01 warmup ratio over 50 rounds. Specifically, the Hateful-Memes dataset ends training after 25 rounds, the CrisisMMD dataset after 40 rounds, and the MedAlpaca and VQA-RAD datasets after 50 rounds. In the benchmark, the results are obtained by splitting with different random seeds and averaging over three runs, reporting the mean and standard deviation. The construction methods of different datasets in various multimodal scenarios are detailed in~\cref{sec:multimodal}. More training details can be found in Suppl.

\begin{table*}[!t]
    \centering
    \renewcommand{\arraystretch}{1.2}
    \setlength{\tabcolsep}{5pt}
    \fontsize{9}{9}\selectfont
    \begin{tabular}{l|l|c|c|c|c|c}
    \hline
    \multirow{2}{*}{MLLMs} & \multirow{2}{*}{Method} & \multirow{2}{*}{Comm. cost}  & \multicolumn{4}{c}{Selected/Total Clients} \\ \cline{4-7}
      & & & 2/10 & 4/10 & 5/20 & 10/30 \\ \hline
    \multirow{3}{*}{MiniCPM-V-2\_6$_{int4}$} & Zero-shot & 0M & 66.57$\pm$0.00 & 66.57$\pm$0.00 & 66.57$\pm$0.00 & 66.57$\pm$0.00 \\
    & Local & 0M & 66.80$\pm$0.11 & 66.80$\pm$0.11 & 64.46$\pm$0.13 & 63.50$\pm$0.08 \\
    & FL & 617M & 75.58$\pm$1.72 & 76.42$\pm$0.34 & 69.70$\pm$0.45 & 66.91$\pm$0.62 \\ \hline
    \multirow{3}{*}{MiniCPM-Llama3-V-2\_5$_{int4}$} & Zero-shot & 0M & 69.29$\pm$0.00 & 69.29$\pm$0.00 & 69.29$\pm$0.00 & 69.29$\pm$0.00 \\
    & Local & 0M & 67.02$\pm$0.09 & 67.02$\pm$0.09 & 63.21$\pm$0.05 & 63.26$\pm$0.10 \\
    & FL & 621M & 72.44$\pm$0.51 & 70.19$\pm$0.64 & 68.82$\pm$0.99 & 71.16$\pm$0.72 \\ \hline
    \end{tabular}
    \caption{Comparison of different lightweight MLLMs based on selected clients per round and total clients (Selected/Total Clients). ``FL" denotes FedAdam. The communication cost of LoRA-based fine-tuning is affordable in cross-silo FL scenarios.}
    \label{tab:mllms}
    \vspace{-5pt}
\end{table*}

\subsection{Performance Evaluation}

\noindent{\textbf{Comparison of different baselines.}} \cref{tab:aligned}-\cref{tab:Medical} sequentially present the results of the ``Zero-shot", ``Local", and FL baselines on Hateful-Memes, CrisisMMD, VQA-RAD, and SLAKE across different levels of modality heterogeneity. In most scenarios, the performance of ``Local" and the majority of FL baselines surpasses that of ``Zero-Shot", indicating that existing MLLMs still require further fine-tuning on private datasets for better real-world deployment. Compared to ``Local", most FL baselines show significant improvement across different scenarios, especially on the multi-class CrisisMMD dataset, where FedAvgM does not perform well in situations with strong modality heterogeneity. In the medical visual question answering task, which includes both open- and closed-ended questions, the improvement of ``Local" and FL over ``Zero-shot" is more pronounced in closed-ended questions. Compared to VQA-RAD, the open-ended questions in the unseen SLAKE dataset pose more challenges for ``Local" and ``FL".

\noindent\textbf{MLLMs, clients number, and comm. cost.} As shown in~\cref{tab:mllms}, we experimented with lightweight MLLMs ``MiniCPM-V-2\_6int4" and ``MiniCPM-Llama3-V-2\_5int" across different client counts. For different MLLMs, the FL baseline still show a clear advantage compared to the ``Zero-shot" and ``Local" baselines. This federated fine-tuning approach ensures data privacy while expanding the data range beyond local data, further boosting the performance of MLLMs. As the number of clients increases, sample size per client decreases, reducing ``Local" performance. However, with 30 clients, the FL baseline with ``V-2\_5" MLLM improves by 1.87 over ``Zero-shot", and with 20 clients, ``V-2\_6" improves by 3.13. This shows the FL baseline can handle client fluctuations in a cross-silo setting, with a affordable communication cost of around 620M.

\begin{figure}[!t]
    \centering
    \includegraphics[width=0.8\linewidth]{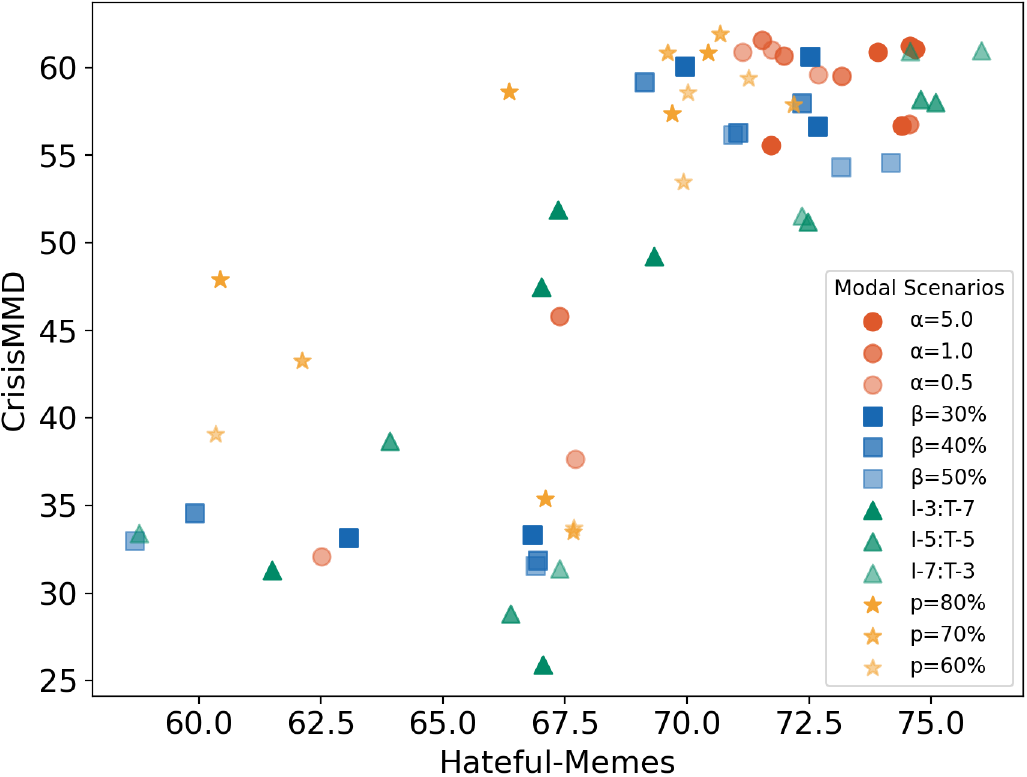}
    \caption{Influence of different modal scenarios. Aligned modal scenario ($\alpha$). Missing modal scenario ($\beta$). Cross modal scenario ($\text{I-}*:\text{T-}*$). Hybrid modal scenario ($p$).}
    \label{fig:scenario}
    \vspace{-5pt}
\end{figure}

\noindent{\textbf{Influence of different multimodal scenarios.}}
\cref{fig:scenario} shows the overall performance differences across different modal scenarios and reveals the influence of these scenarios on performance. We can see that: 1) Each modal scenario contains certain heterogeneous conditions that significantly degrade the performance of FedMLLM on both datasets.
2) Hybrid modal scenario outperforms other non-aligned modal scenarios in overall performance on the CrisisMMD dataset. However, on the Hateful-Memes dataset, its best performance is generally lower than that of other non-aligned modal scenarios.  
3) The performance distributions of the Missing and Hybrid modal scenarios exhibit similar trends, whereas the Cross modal scenario has the widest performance distribution, encompassing both the best performance on the Hateful-Memes dataset and the worst performance on the CrisisMMD dataset.

\begin{table}[!t]
\renewcommand{\arraystretch}{1.2}
    \centering
    \setlength{\tabcolsep}{8.5pt}
    \fontsize{9}{9}\selectfont
    \begin{tabular}{l|cc}
         \hline
         Scenario & Hateful-Memes & CrisisMMD \\ \hline
         \multirow{2}{*}{Missing} & 70.94 & 56.17 \\
         & 71.66$_{(\color{highlightcolor}+0.72)}$ & 57.20$_{(\color{highlightcolor}+1.03)}$ \\ \hline
         \multirow{2}{*}{Cross} & 69.33 & 49.22 \\
         & 71.17$_{(\color{highlightcolor}+1.84)}$ & 49.56$_{(\color{highlightcolor}+0.34)}$ \\ \hline
         \multirow{2}{*}{Hybrid} & 69.69 & 53.45 \\
         & 70.59$_{(\color{highlightcolor}+0.90)}$ & 58.46$_{(\color{highlightcolor}+5.01)}$ \\ \hline
    \end{tabular}
    \caption{Results of modality-agnostic strategies in different scenarios. The original results are from FedAdagrad, which is most affected by multimodal heterogeneity, using its most detrimental non-aligned modality configurations.}
    \label{tab:strategy}
\end{table}

\begin{figure}[!t]
    \centering
    \includegraphics[width=1.0\linewidth]{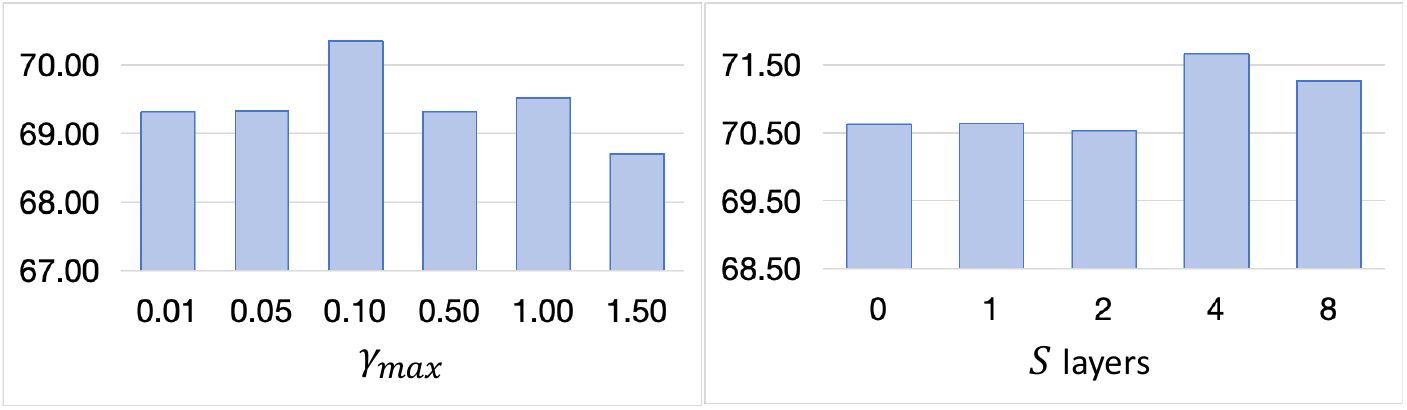}
    \caption{Performance of different values for $\gamma_{\text{max}}$ and $S$ layers, with the horizontal axis representing $\gamma_{\text{max}}$ (left) and $S$ (right), and the vertical axis showing results on the Hateful Memes dataset.}
    \label{fig:y_max_S}
    \vspace{-5pt}
\end{figure}

\noindent{\textbf{Results on modality-agnostic strategies.}} \cref{tab:strategy} and \cref{tab:strategy-fl} demonstrate the performance improvement of two modality-agnostic strategies across various non-aligned modal scenarios and FL methods, respectively.
The listed non-aligned scenarios are those in which performance is most negatively affected, while the addition of both modality-agnostic strategies improves performance in these scenarios. For different FL methods, both strategies also help mitigate the negative effects of modal heterogeneity. Specifically, FedAdagrad, which is significantly negatively affected by modality heterogeneity, improves with both strategies, showing a 5.01\% gain on the CrisisMMD dataset in the Hybrid modal scenario. 

Furthermore, \cref{tab:strategy_ablation} presents ablation studies showing that prompt and regularization strategies effectively mitigate modality heterogeneity, yielding the best performance when applied together.
Compared to FedProx~\cite{li2020federated}, which applies fixed regularization across all layers, our approach dynamically adjusts regularization strength based on modality heterogeneity, achieving superior performance in complex scenarios.
\cref{fig:y_max_S} shows the ablation study for the $\gamma_{\text{max}}$ and $S$ layers in improved regularization, with final values of $\gamma_{\text{max}}$ and $S$ set to 0.1 and 4, respectively.

\begin{table}[!t]
\renewcommand{\arraystretch}{1.2}
    \centering
    \setlength{\tabcolsep}{8.5pt}
    \fontsize{9}{9}\selectfont
    \begin{tabular}{l|cc}
    \hline
       Method  & Hateful-Memes & CrisisMMD \\ \hline
       \multirow{2}{*}{FedYogi} & 67.02 & 47.46 \\
       & 69.80$_{(\color{highlightcolor}+2.78)}$ & 49.46$_{(\color{highlightcolor}+2.00)}$ \\ \hline
       \multirow{2}{*}{FedAdam} & 67.35 & 51.88 \\
       & 69.00$_{(\color{highlightcolor}+1.65)}$ & 56.03$_{(\color{highlightcolor}+4.15)}$ \\ \hline
       \multirow{2}{*}{FedAvgM} & 61.50 & 31.31 \\
       & 62.97$_{(\color{highlightcolor}+1.47)}$ & 33.13$_{(\color{highlightcolor}+1.82)}$ \\ \hline
       \multirow{2}{*}{FedAdagrad} & 69.33 & 49.22 \\
       & 71.17$_{(\color{highlightcolor}+1.84)}$ & 49.56$_{(\color{highlightcolor}+0.34)}$  \\ \hline
    \end{tabular}
    \caption{Results of modality-agnostic strategies in different FL methods. The original results are from the most detrimental cross modal scenario ($\text{I-}3:\text{T-}7$).}
    \label{tab:strategy-fl}
\end{table}

\begin{table}[!t]
\renewcommand{\arraystretch}{1.2}
    \centering
    \setlength{\tabcolsep}{3.5pt}
    \fontsize{9}{9}\selectfont
    \begin{tabular}{c|c|c|c|c}
    \hline
        $+$Prompt & $+$Reg. & $+$FedProx & Hateful-Memes & CrisisMMD \\ \hline
        & & & 69.93 & 53.45 \\
        \checkmark & & & 73.51$_{(\color{highlightcolor}+3.58)}$ & 53.86$_{(\color{highlightcolor}+0.41)}$ \\
         & \checkmark & & 73.41$_{(\color{highlightcolor}+3.48)}$ & 53.82$_{(\color{highlightcolor}+0.37)}$ \\
        \checkmark & & \checkmark & 72.31$_{(\color{highlightcolor}+2.38)}$ & 55.68$_{(\color{highlightcolor}+2.23)}$ \\
        \checkmark & \checkmark & &
        73.69$_{(\color{highlightcolor}+3.76)}$ & 58.46$_{(\color{highlightcolor}+5.01)}$ \\ \hline
    \end{tabular}
    \caption{Ablation study of modality-agnostic strategies for FedAdagrad in the Hybrid Modal Scenario with $p=60\%$.}
    \label{tab:strategy_ablation}
    \vspace{-10pt}
\end{table}

\section{Discussion}
\noindent\textbf{Number of modality.} FedMLLM currently focuses on the most fundamental and essential image-text multimodal data. However, other modalities—such as video and audio—still require further exploration in future work. When introducing video modalities, especially long videos, a key challenge is achieving efficient training and deployment given the limited computational resources on clients.

\noindent\textbf{Splitting of data.} The data partitioning methods for the four modality scenarios are widely used and fundamental, offering researchers a simple and efficient environment for data construction and testing. However, real-world environments are much more complex, and we look forward to more benchmark studies in the future that provide more diverse and challenging multimodal scenarios.

\noindent\textbf{Number of client.} We focus on the cross-silo FL configuration, typically involving small institutions like hospitals with fewer than 100 clients. In our experiments, we evaluated the impact of client numbers on FL performance using 10, 20, and 30 clients, with different clients randomly selected in each round. Cross-device FL, involving thousands of resource-constrained devices like smartphones, presents significant challenges for federated fine-tuning of MLLMs and may be a future research direction.

\noindent\textbf{Diversity of data.} Our benchmark includes five datasets from social media, natural disasters, and medical knowledge domains, covering classification and VQA tasks. However, FedMLLM is not limited to these datasets. With the scene construction descriptions, evaluation details, and data templates provided in the main text and supplementary materials, researchers can adapt it to their own datasets for training and testing.

\noindent\textbf{Adequacy of assessment.} In this benchmark, we evaluated four scenarios, including zero-shot performance of the untuned MLLM, comparisons with the Local baseline and four classic FL methods, and AUC and F1 metrics for classification tasks. Two lightweight MLLMs were tested under different client configurations. For the Med-VQA task, accuracy on open-end and closed-end questions was assessed by GPT-4. Additionally, the unseen SLAKE dataset was evaluated, with open-end questions presenting a particular challenge. Future research in this area is anticipated.

\noindent\textbf{Efficiency analysis.} Our method enables the adaptation of general MLLMs to downstream tasks, such as hateful meme classification in social media, rescue-type classification in natural disasters, and medical visual question answering based on radiology images and questions, via federated learning with an affordable communication cost. Optimizing efficiency will be a focus of future work.

\section{Conclusion}
In this work, we presented a multimodal heterogeneity benchmark to evaluate the performance of federated fine-tuning for MLLMs across six FL baselines on five downstream datasets, incorporating more than ten multimodal heterogeneities across four multimodal scenarios. To tackle the challenges posed by complex multimodal heterogeneities, we have also designed two simple yet effective modality-agnostic strategies. In the proposed FedMLLM framework, we conduct a series of experiments to highlight the advantages of FL over local training, examine the impact of multimodal heterogeneities, and evaluate the effectiveness of the two modality-agnostic strategies.

{
    \small
    \bibliographystyle{ieeenat_fullname}
    \bibliography{main}
}

\clearpage
\setcounter{page}{1}
\maketitlesupplementary

\setcounter{section}{0}


\noindent Due to space limitations, this supplementary material provides further details and clarifications, as follows,
\begin{enumerate}[itemsep=0pt]
    \item Related Work
    \begin{enumerate}[label=\alph*)]
        \item Multimodal Large Language Models.
        \item Multimodal Federated Learning.
        \item Federated Learning.
    \end{enumerate}
    \item Dataset in details.
    \item More visualizations of multimodal scenarios.
    \item More training details.
    \begin{enumerate}[label=\alph*)]
        \item Configurations.
        \item Gradient Norm and Loss Curves.
        \item Complete Formatting Instructions.
    \end{enumerate}
\end{enumerate}

\section{Related Work}
\noindent\textbf{Multimodal Large Language Models.} 
The rapid development of Large Language Models (LLMs) has significantly driven the research surge in Multimodal Large Language Models (MLLMs)~\cite{yin2023survey}. Notably, the emergence of commercial models such as GPT-4~\cite{openai2022chatgpt} and Gemini~\cite{team2023gemini} has attracted widespread attention. Simultaneously, increasingly powerful open-source models~\cite{li2023blip, chiang2023vicuna, zhu2023minigpt} have also emerged. BLIP-2~\cite{li2023blip} is well-known for its Q-former module, with promising VQA results. Building on BLIP-2 and Vicuna~\cite{chiang2023vicuna}, MiniGPT-4~\cite{zhu2023minigpt}, derived from GPT-4~\cite{openai2022chatgpt}, is proposed for image-text pairs. Unlike previous methods, LLaVA~\cite{liu2023visual} is trained on multimodal instruction data generated by GPT-4 to enhance its zero-shot capability. 

However, the large number of parameters in MLLMs leads to high computational costs for training and deployment, limiting their broader application. Recently, efficient MLLMs have also gained significant attention within the academia~\cite{jin2024efficient}, with representative models including Phi~\cite{javaheripi2023phi}, Gemma~\cite{banks2024gemma}, MobileLLM~\cite{liu2024mobilellm}, and MiniCPM-V~\cite{yao2024minicpm}. Compared to previous approaches, MiniCPM-V~\cite{yao2024minicpm} optimizes architecture, training, and inference to achieve superior performance with fewer parameters. Consequently, we utilize MiniCPM-V as the MLLM for both the client and server in the federated learning process.

\noindent\textbf{Multimodal Federated Learning.} 
Unlike the more established single-modal FL, Multimodal Federated Learning (MFL) has significant potential for exploration~\cite{feng2023fedmultimodal}. Existing MFL methods can also be categorized based on the different modal scenarios in our benchmark.
In aligned modal scenarios, MFL techniques resemble those of single-modal FL, focusing on client-specific and shared knowledge~\cite{chen2024feddat}. Some methods also draw from multimodal learning, especially in feature fusion~\cite{feng2023fedmultimodal}. In missing modal scenarios, some MFL methods rely on pre-trained generative models to fill the gaps in modality~\cite{che2024leveraging}. In cross modal scenarios, some MFL methods employ a two-branch structure to separately constrain the learning of specific and shared modalities~\cite{yang2024cross,yang2022cross,qi2024adaptive}, while hybrid modal MFL methods utilize various graph techniques to capture complex correlations among clients~\cite{chen2022fedmsplit,chen2024disentanglement}.

Typically, MFL extracts features from raw inputs using pre-trained models before training, but its performance is ultimately limited by the effectiveness of these models in feature extraction~\cite{feng2023fedmultimodal,yang2024cross}. With the rise of MLLMs, their application in MFL has become a trend. However, only one study so far has attempted to combine MLLMs with FL on the server side to address issues of data heterogeneity and long-tail distribution~\cite{zhang2024mllm}. In this paper, we present the first benchmark for federated fine-tuning of MLLMs, targeting diverse multimodal heterogeneities.

\noindent\textbf{Federated Learning.}
As a classic framework for privacy protection, Federated Learning (FL) typically allows multiple parties to co-train while keeping the data local~\cite{huang2024federated}. The original FedAvg~\cite{mcmahan2017communication} method aggregates model weights from different clients solely based on sample proportions, which makes it challenging to effectively address complex data heterogeneity. As a result, various variants have emerged: some constrain the update direction of the local model on the client side~\cite{li2020federated, karimireddy2020scaffold}, while others optimize the aggregation method on the server side~\cite{reddiadaptive, hsu2019measuring, rehman2023dawa}. 

Our benchmark includes four representative FL methods for server-side aggregation. FedYogi, FedAdam, and FedAdagrad~\cite{reddiadaptive} apply the centralized Yogi~\cite{zaheer2018adaptive}, Adam~\cite{kingma2014adam}, and Adagrad~\cite{duchi2011adaptive} optimizers to the server-side aggregation in the FL framework, while FedAvgm~\cite{hsu2019measuring} uses momentum updates on the server, similar to momentum applied on top of SGD. Because the original FedAvg~\cite{mcmahan2017communication} and FedProx~\cite{li2020federated} with a global regularization term face training instability during federated fine-tuning of MLLM, the conventional average aggregation method does not converge effectively. As a result, these two methods are not included separately in the comparison baselines.

\begin{figure*}[!t]
    \centering
    \includegraphics[width=1.0\linewidth]{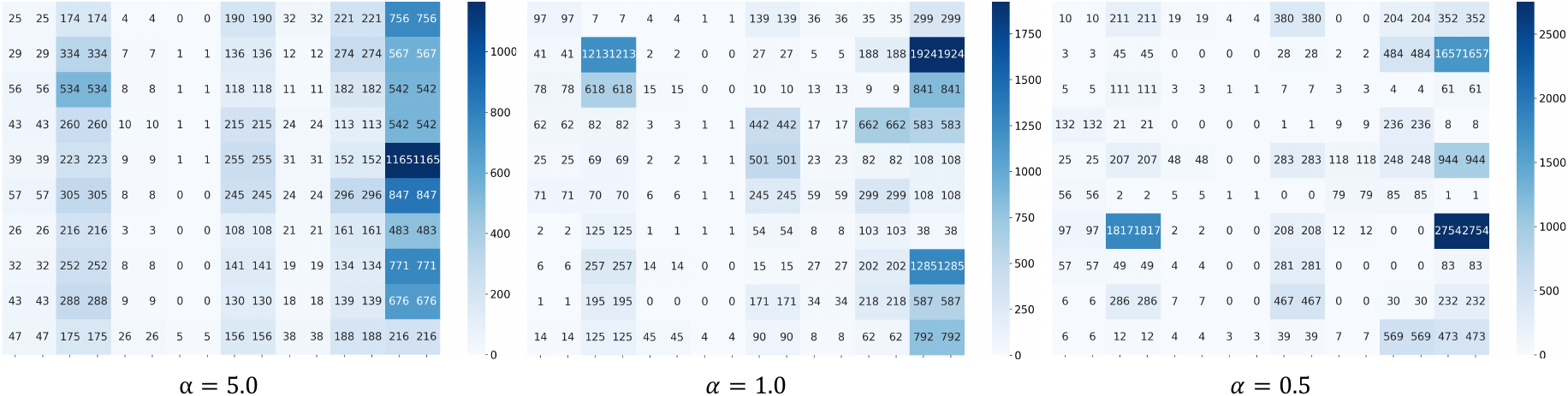}
    \caption{Visualization of heterogeneity in the aligned modal scenario on the CrisisMMD dataset. Each row represents a client dataset, and each pair of columns represents a class with both image and text modalities. The values inside represent the sample count.}
    \label{fig:aligned_modal}
\end{figure*}

\begin{figure*}[!t]
    \centering
    \includegraphics[width=1.0\linewidth]{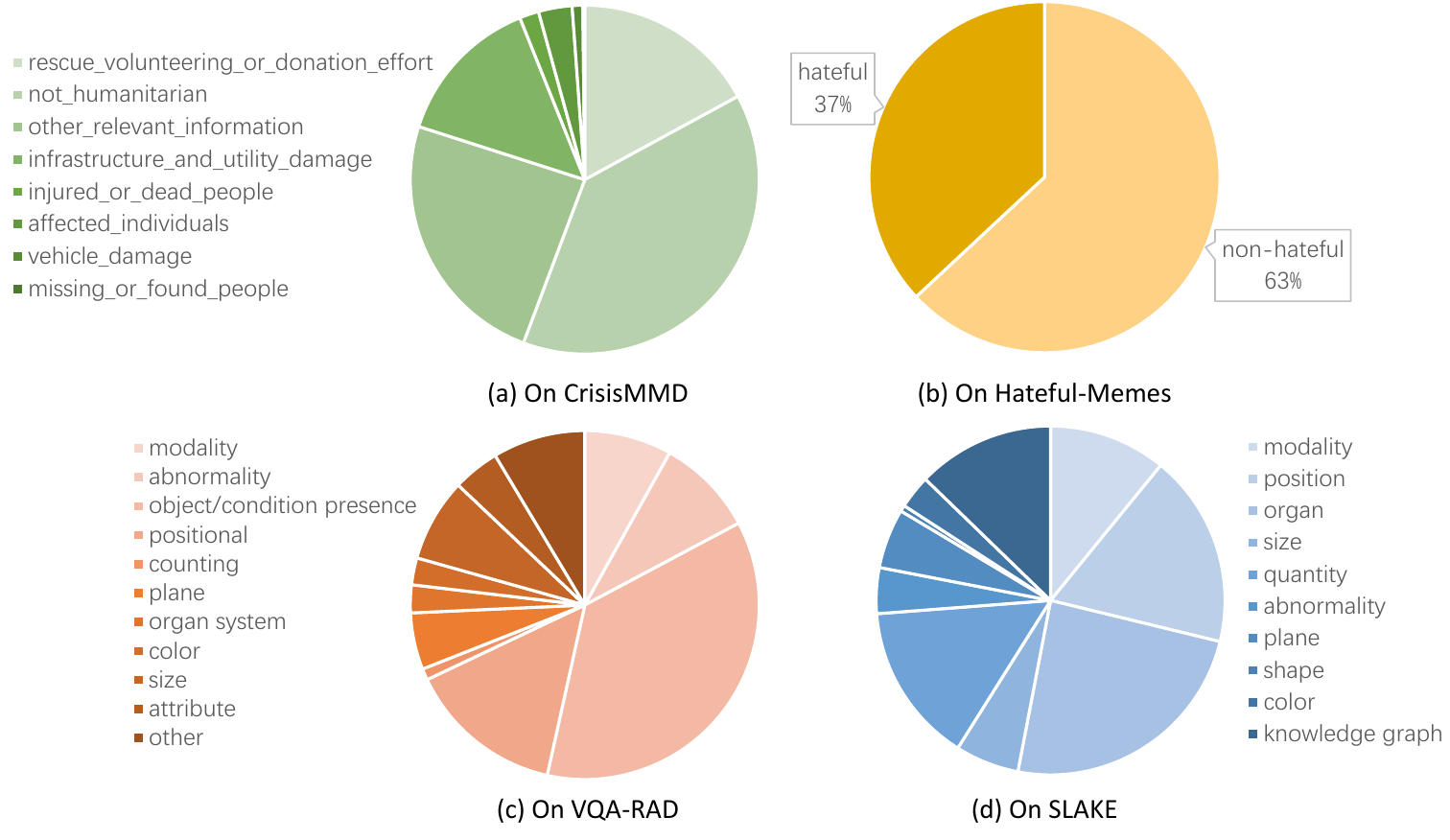}
    \caption{Visualization of category and question types across the CrisisMMD, Hateful-Memes, VQA-RAD, and SLAKE datasets.}
    \label{fig:task_class_type}
\end{figure*}

\begin{figure*}[!t]
    \centering
    \includegraphics[width=0.95\linewidth]{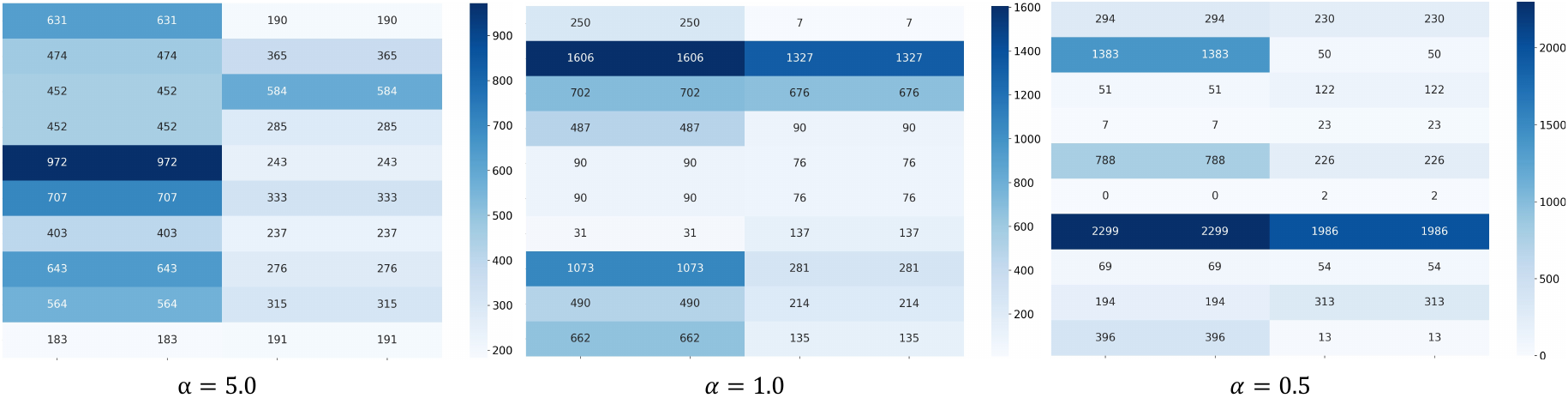}
    \caption{Visualization of heterogeneity in the aligned modal scenario on the Hateful-Memes dataset. Each row represents a client dataset, and each pair of columns represents a class with both image and text modalities. The value inside represent the sample count.}
    \label{fig:aligned_hateful}
\end{figure*}

\begin{figure*}[!t]
    \centering
    \includegraphics[width=0.95\linewidth]{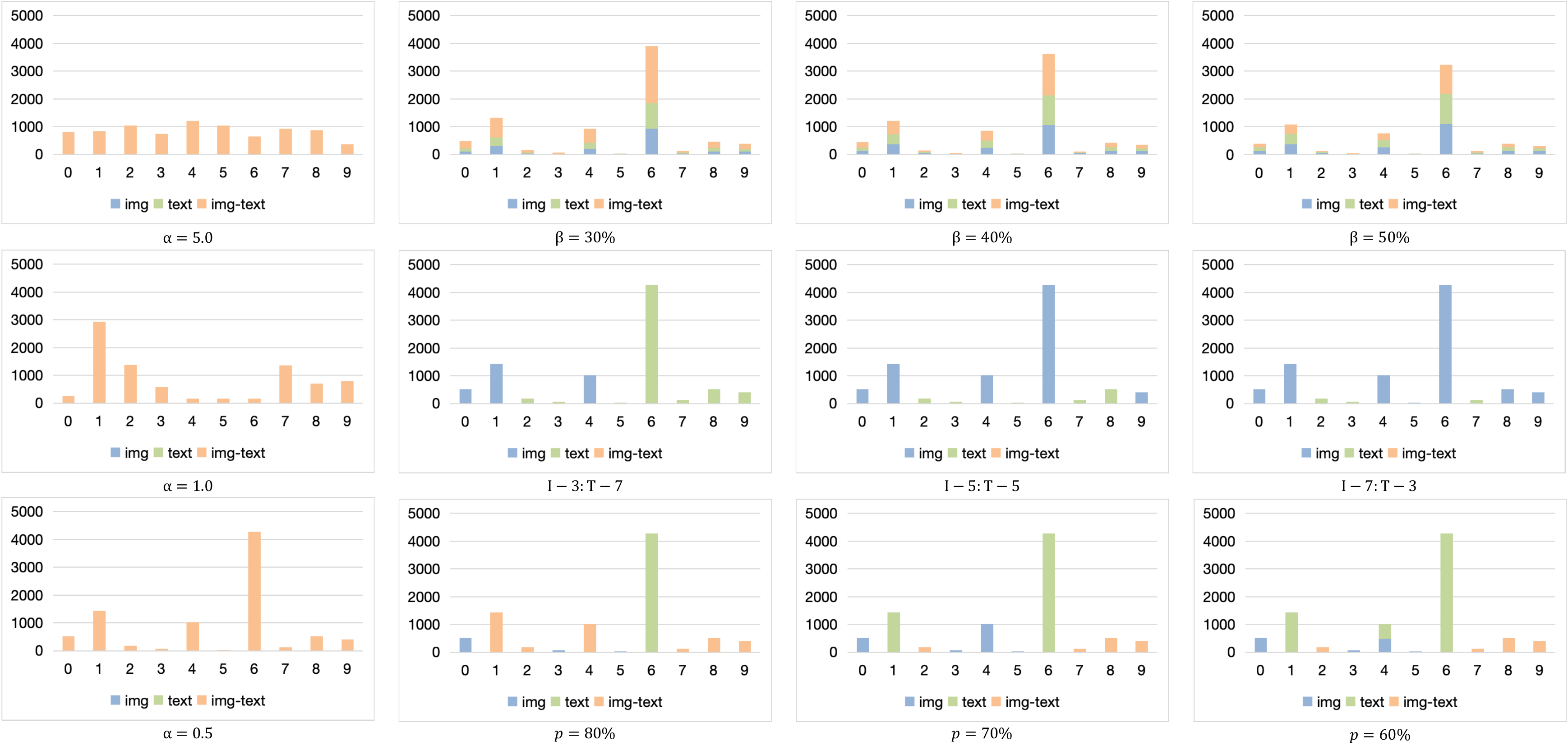}
    \caption{Visualization of multimodal heterogeneity across four multimodal scenarios on the Hateful-Memes dataset. The horizontal axis represents each client, and the vertical axis indicates the sample count.}
    \label{fig:modal-heter-hateful}
\end{figure*}

\begin{figure}
    \centering
    \includegraphics[width=0.65\linewidth]{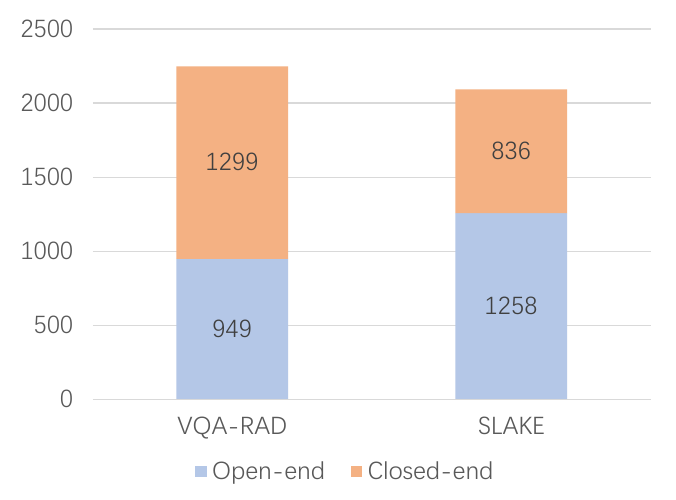}
    \caption{Visualization of open-ended and closed-ended answer counts on the VQA-RAD and SLAKE datasets.}
    \label{fig:open-closed}
    \vspace{-10pt}
\end{figure}

\section{Dataset}
\noindent\textbf{Hateful-Memes~\cite{kiela2020hateful}} contains a total of 10k memes  focused on detecting hate speech in multimodal memes, with the training set, dev set, and test set comprising 85\%, 5\%, and 10\%, respectively. In this study, we only use the training set and the test set. The dataset includes a total of five types of data: multimodal hate, unimodal hate, benign image, benign text, and random not-hateful data. The task aims to determine if the given image and text are hateful, a binary classification problem. The evaluation metric used is the Area Under the Receiver Operating Characteristic Curve (ROC AUC), as recommended by the authors.

\noindent\textbf{CrisisMMD~\cite{alam2018crisismmd}} includes 18.1k tweets from major natural disasters that occurred in 2017 across different parts of the world. We use a training set of 13.6k tweets and a test set of 2.2k tweets, excluding the dev set. The dataset comprises eight humanitarian categories for classification: affected individuals, infrastructure and utility damage, injured or dead people, missing or found people, rescue volunteering or donation effort, vehicle damage, other relevant information, not humanitarian. The evaluation metric is the F1 score~\cite{feng2023fedmultimodal}.

\noindent\textbf{MedAlpaca~\cite{han2023medalpaca}} contains approximately 34k Q\&A pairs derived from the front and back of medical flashcards, with images removed to retain only concise and targeted questions and answers. This dataset covers areas such as anatomy, physiology, pathology, and pharmacology, aiming to support the learning and understanding of basic medical sciences, clinical knowledge, and clinical skills.

\noindent\textbf{VQA-RAD~\cite{lau2018dataset}} is a dataset for Medical Visual Question Answering (Med-VQA), consisting of 2,248 Q\&A pairs and 315 radiology images. The training set includes 1,797 Q\&A pairs, while the test set contains 451 Q\&A pairs. This Med-VQA dataset includes 11 question types: modality, abnormality, and more.
The evaluation metric follows the GPT-4~\cite{achiam2023gpt} accuracy evaluation for open-ended and closed-ended answers, as mentioned in~\cite{liu2024pefomed}.

\noindent\textbf{SLAKE~\cite{liu2021slake}} is an English-Chinese bilingual dataset for the Med-VQA task, containing 14k Q\&A pairs and 642 images with 2,070 test pairs. The questions in this Med-VQA dataset include vision-only and knowledge-based types. According to~\cite{liu2024pefomed}, the evaluation metrics include the accuracy of open-ended and closed-end answers, as well as overall accuracy, assessed by GPT-4~\cite{achiam2023gpt}.

\section{Visualization}

\noindent\textbf{Modality heterogeneity.} In the main text, we show the visualizations of the heterogeneity of aligned modal scenarios for the CrisisMMD dataset and the multimodal heterogeneity visualizations for the four multimodal scenarios in the Hateful-Memes dataset, as shown in~\cref{fig:aligned_modal} and~\cref{fig:non-aligned_modal}. Furthermore, \cref{fig:aligned_hateful} and~\cref{fig:modal-heter-hateful} supplement the visualizations of aligned modal heterogeneity and provide additional multimodal heterogeneity visualizations for the four multimodal scenarios in the Hateful-Memes dataset. Since the multimodal heterogeneity in the CrisisMMD dataset is similar to that in the Hateful-Memes dataset, redundant visualizations are omitted here for brevity.

As shown in~\cref{fig:aligned_modal} and~\cref{fig:aligned_hateful}, each dataset in our benchmark includes three levels of data heterogeneity. As the $\alpha$ value decreases, the differences in data categories and quantities among clients increase. The visualizations of multimodal heterogeneities are presented in~\cref{fig:non-aligned_modal} and~\cref{fig:modal-heter-hateful}. In the aligned modal scenario, each client contains samples with all modalities. Based on data heterogeneity with $\alpha=0.5$, we further construct scenarios with multimodal heterogeneity under different non-aligned modal conditions. In the missing modal scenario, varying missing rates result in different proportions of missing modalities in the samples, although each client’s dataset still contains samples with all modalities. In the cross modal scenario, the proportion of single-modal clients varies. In the hybrid modal scenario, clients consist of both single-modal and multimodal combinations.

In addition to the category heterogeneity in Hateful-Memes and CrisisMMD, we also visualize the heterogeneity brought by different category types and question types in the datasets. As shown in~\cref{fig:task_class_type}, CrisisMMD and Hateful-Memes each contain 8 rescue categories and a binary classification for malicious memes; VQA-RAD and SLAKE are two Med-VQA datasets, with 11 and 10 question categories, respectively, and there are differences in the quantities. Additionally, these two Med-VQA datasets include both open-ended and closed-ended answer types, with the quantity differences shown in~\cref{fig:open-closed}.

\section{Training Details}

\noindent\textbf{Configurations.} The local baseline uses 5 training epochs for optimal performance, as shown in~\cref{tab:epoch}. The FL baselines train for 50 rounds, with 1 epoch per round. The difference in the number of epochs is normal~\cite{ye2024openfedllm,ye2025fedllm}, as the local baseline aims to achieve the best result, while training too many epochs per round in the FL baselines may introduce more bias.
\begin{table}[!h]
    \renewcommand{\arraystretch}{1.1}
    \centering
    \setlength{\tabcolsep}{7pt}
    \fontsize{9}{9}\selectfont
    \begin{tabular}{@{}c|c|c|c|c|c|c@{}}
        \hline
        Ep & 1 & 2 & 3 & 4 & 5 & 6 \\ \hline
        C & 25.22 & 26.61 & 32.85 & 31.44 & 33.34 & 32.06 \\ 
        H & 65.74 & 65.76 & 65.86 & 66.05 & 66.83 & 66.44 \\ \hline
    \end{tabular}
    \caption{Comparison of different training epochs on Hateful-Memes (H) and CrisisMMD (C).}
    \label{tab:epoch}
    \vspace{-5pt}
\end{table}
We train on a lightweight MLLM using LoRA, with LoRA parameters transmitted between the client and server, which is acceptable in FL. In the benchmark, the FL baselines differ only in the aggregation computation on the server side, with minimal impact on efficiency. As shown in~\cref{tab:time}, we provide the communication cost, computational cost, and total training time for Hateful-Memes (H) and CrisisMMD (C).
\begin{table}[!h]
    \renewcommand{\arraystretch}{1.2}
    \centering
    \setlength{\tabcolsep}{8pt}
    \fontsize{9}{9}\selectfont
    \begin{tabular}{@{}c|c|c|c@{}}
        \hline
        Trainable / Total & Comm. cost & \multicolumn{2}{c}{Train time} \\ \hline
        617M / 8716M & 617M & H: 11h 05m & C: 21h 04m \\ \hline
    \end{tabular}
    \caption{The communication cost, computational cost, and total training time of Hateful-Memes and CrisisMMD.}
    \label{tab:time}
    \vspace{-5pt}
\end{table}

In cross modal scenario, suppose $\gamma$ is set to $\gamma_{\text{max}}$. When the missing rate is $\beta\%$ in missing modal scenario, $\gamma$ is linearly adjusted to 
$\gamma_{\text{max}} \times \beta\%$
to adaptively adjust the strength of agreement with the global model. The hybrid modal scenario combines aligned, missing, and cross modal scenarios. Thus, the $\gamma$ setting follows these scenarios, with no regularization in the aligned case. In practice, $\beta\%$ is calculated from the client's private dataset.

\begin{figure}[!t]
    \centering
    \begin{subfigure}[b]{0.45\textwidth}
        \includegraphics[width=\textwidth]{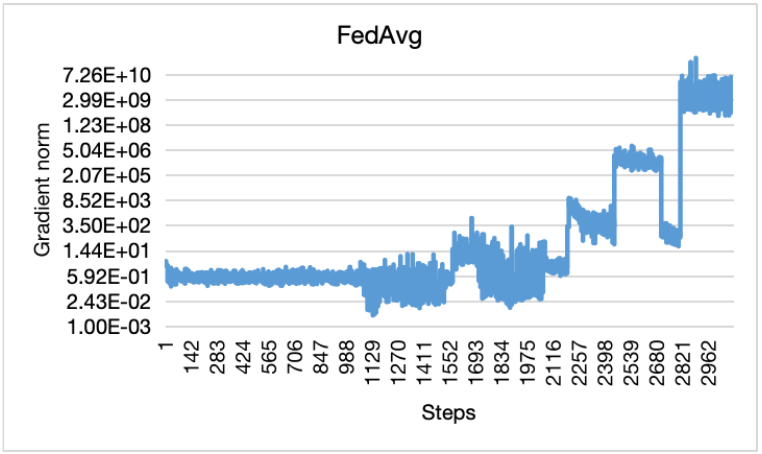}
        \caption{The gradient norm curve.}
        \label{fig:subfig1}
    \end{subfigure}
    \hfill
    \begin{subfigure}[b]{0.45\textwidth}
        \includegraphics[width=\textwidth]{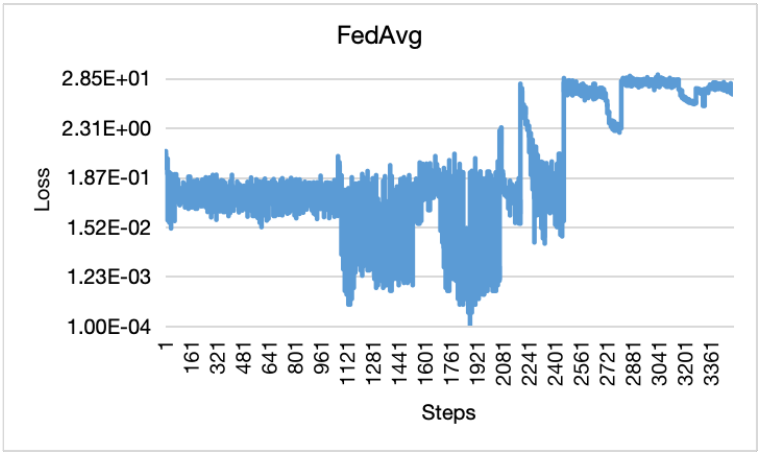}
        \caption{The loss curve.}
        \label{fig:subfig2}
    \end{subfigure}
    \caption{The gradient norm and loss curves for FedAvg.}
    \label{fig:fedavg}
    \vspace{-10pt}
\end{figure}

\begin{figure}[!t]
    \centering
    \begin{subfigure}[b]{0.45\textwidth}
        \includegraphics[width=\textwidth]{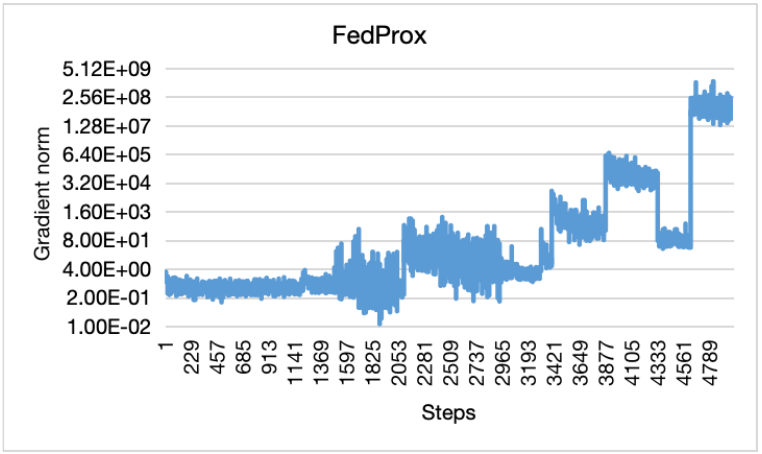}
        \caption{The gradient norm curve.}
        \label{fig:subfig3}
    \end{subfigure}
     \hfill
    \begin{subfigure}[b]{0.45\textwidth}
        \includegraphics[width=\textwidth]{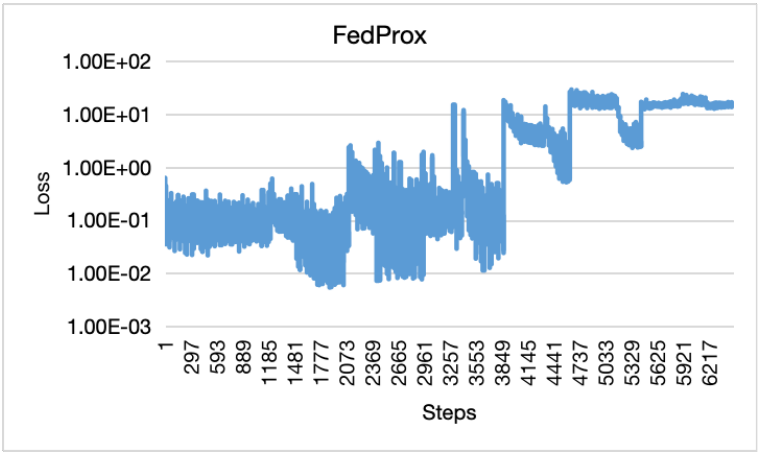}
        \caption{The loss curve.}
        \label{fig:subfig4}
    \end{subfigure}
    \caption{The gradient norm and loss curves for FedProx.}
    \label{fig:fedprox}
    \vspace{-10pt}
\end{figure}

\noindent\textbf{Gradient Norm and Loss Curves.} FedAvg~\cite{mcmahan2017communication} and FedProx~\cite{li2020federated} methods are used to train the FedMLLM framework. However, due to the inability of average aggregation to effectively handle the impact of heterogeneity, LoRA experiences unstable training, causing both the loss and gradient norm to gradually increase, thereby hindering the effective convergence of the global model. In~\cref{fig:fedavg} and~\cref{fig:fedprox}, the gradient norms and loss curves for both FedAvg and FedProx show a gradual upward trend. While the regularization term in FedProx helps partially mitigate this increase, it ultimately falls short in addressing the challenges posed by heterogeneity in the training of FedMLLM.

\begin{table*}[!t]
\centering
\begin{tcolorbox}[colback=gray!20, colframe=black, width=\textwidth, rounded corners, boxrule=0.3mm]
``\textbf{id}": \{id\}, \\
``\textbf{image}": \{image\}, \\
``\textbf{conversations}": \\
\{``\textbf{role}": ``user",
``\textbf{content}": ``\textless image\textgreater\textbackslash nSelect the best answer to the following multiple-choice question, without considering the modality.\textbackslash n\{question\}, without considering the modality?\textbackslash nOptions:\{options\} Answer with the option's letter from the given choices directly and only give the best option. The best answer is:"\}, \\
\{``\textbf{role}": ``assistant",
``\textbf{content}": \{class\}\}
\end{tcolorbox}
\caption{Prompt template for FedMLLM.}
\label{tab:com_instruct}
\vspace{-10pt}
\end{table*}

\begin{table*}[!t]
\centering
\begin{tcolorbox}[colback=gray!20, colframe=black, width=\textwidth, rounded corners, boxrule=0.3mm]
\textit{Original Prompt}\\
\textbf{Question:} Is the content hateful {\color{highlightcolor}based on the text and image}?\\
\textbf{Instruction:} \textless image\textgreater\textbackslash
n Select the best answer to the following multiple-choice question {\color{highlightcolor}based on the text and image}.\textbackslash
n\{text\}\textbackslash
n\{question\}\textbackslash
nOptions:\textbackslash
n(A) not-hateful\textbackslash
n(B) hateful\textbackslash
nAnswer with the option's letter from the given choices directly and only give the best option. The best answer is: \\

\textit{Augmented Prompt}\\
\textbf{Question:} Is the content hateful, {\color{cyan}without considering the modality}?\\
\textbf{Instruction:} \textless image\textgreater\textbackslash
n Select the best answer to the following multiple-choice question, {\color{cyan}without considering the modality}.\textbackslash
n\{text\}\textbackslash
n\{question\}\textbackslash
nOptions:\textbackslash
n(A) not-hateful\textbackslash
n(B) hateful\textbackslash
nAnswer with the option's letter from the given choices directly and only give the best option. The best answer is:
\end{tcolorbox}
\caption{An example of the modal-agnostic prompt strategy applied to the Hateful-Memes dataset.}
\vspace{-10pt}
\label{fig:org-aug}
\end{table*}

\noindent\textbf{Complete Formatting Instructions.} To prepare the fine-tuning data, the data format must be adjusted according to the prompt template shown in~\cref{tab:com_instruct}. Each sample should be structured as a dictionary containing the keys: ``id", ``image", and ``conversations". The ``image" key specifies the path to the image, while ``conversations" includes two dictionaries: one representing the user’s instruction, which contains the question and related information, and the other representing the assistant's (MLLM) response. Since the task primarily involves classification, the response consists of the option letter and the corresponding category. For more details on data construction, please refer to the code provided in the supplementary materials.

\end{document}